\documentclass[10pt]{article}
\usepackage[letterpaper]{geometry}
\usepackage{hicss}
\usepackage{times}
\usepackage[none]{hyphenat}
\usepackage{url}
\usepackage{latexsym}
\usepackage{graphicx}

\usepackage{hyperref}
\usepackage{amsmath}
\usepackage{amssymb}
\usepackage{mathtools}
\usepackage{float}

\title{A2Log: Attentive Augmented Log Anomaly Detection}

\author{
Thorsten Wittkopp$\empty^1$, Alexander Acker$\empty^1$, Sasho Nedelkoski$\empty^1$, Jasmin Bogatinovski$\empty^1$, \\ Dominik Scheinert$\empty^1$, Wu Fan$\empty^2$ and Odej Kao$\empty^1$\\
$\empty^1$Technische Universität Berlin, Germany, \{t.wittkopp, nedelkoski, firstname.lastname\}@tu-berlin.de\\
$\empty^2$Huawei, China, \{wufan3@huawei.com\}
}

\begin{document}
\maketitle

\begin{abstract}
Anomaly detection becomes increasingly important for the dependability and serviceability of IT services.
As log lines record events during the execution of IT services, they are a primary source for diagnostics.
Thereby, unsupervised methods provide a significant benefit since not all anomalies can be known at training time.
Existing unsupervised methods need anomaly examples to obtain a suitable decision boundary required for the anomaly detection task.
This requirement poses practical limitations. 
Therefore, we develop A2Log, which is an unsupervised anomaly detection method consisting of two steps: Anomaly scoring and anomaly decision. 
First, we utilize a self-attention neural network to perform the scoring for each log message. 
Second, we set the decision boundary based on data augmentation of the available normal training data.
The method is evaluated on three publicly available datasets and one industry dataset. 
We show that our approach outperforms existing methods. 
Furthermore, we utilize available anomaly examples to set optimal decision boundaries to acquire strong baselines. 
We show that our approach, which determines decision boundaries without utilizing anomaly examples, can reach scores of the strong baselines. 
\end{abstract}


\section{Introduction}
\label{sec:introduction}
IT systems are rapidly evolving to meet the growing demand for new services and applications in various economic fields. Many companies outsource their services to the cloud ~\cite{rosendo2018improve}. This outsourcing entails growing data centers, increasingly large networks, and interconnected devices to provide the required IT services. Despite accelerating innovations and business opportunities, this trend increases complexity and thus, aggravates the operation and maintenance of these services and systems ~\cite{santos2017analyzing}. Operators of the services require assistance in maintaining control of this complexity to ensure dependability, stability, and serviceability. 
The field of artificial intelligence for IT operations (AIOps) is intended to support service and system operators to meet these challenges.
Thereby, anomaly detection is a vital part, which is applied to monitoring data such as metrics, logs, or traces.

Log data is a primary source for troubleshooting since they record events during the execution of service applications. These logging events evolve due to software updates.
Therefore unsupervised methods are valuable due to their sense of new and unknown anomalies ~\cite{baier2019cope,chandola2009anomaly}. 
The second benefit of unsupervised methods is that they do not require labeled data, which is hard to obtain and cost-intensive ~\cite{wittkopp2020decentralized}. Due to the complexity of IT services, the log data volume is growing to the extent that it cannot be manually analyzed.
Therefore, recent research uses deep learning to analyze log data and perform anomaly detection ~\cite{guo2021logbert, du2017deeplog, nedelkoski2020self}. 
These studies mostly assume the existence of a sufficient amount of labeled validation data for parameter tuning. 
However, in production settings, where services evolve, such data is hard to obtain, volatile, and requires manual evaluation by experts. 
In particular, when it comes to deriving an anomaly decision, these methods lack the capability to obtain a decision in an unsupervised manner. The anomaly decision in all mentioned methods is based on a decision boundary that decides on the respective binary class. Commonly employed methods need to be aware of the anomalies in the validation data to set this decision boundary optimally. This relatively strong requirement poses limitations. 

In this paper, we address these problems via a two-fold solution called \textit{A2Log}. 
First, we create a neural network based on the self-attention mechanism to perform an anomaly scoring. 
Second, we perform data augmentation to generate deviations on the respective training data~\cite{liu2020survey,shorten2019survey}, to analyze the model response and calculate the final decision boundary.
In both steps, only normal training data is used, thus we call our approach unsupervised.

The contributions of this work contain the following:
\begin{itemize}
    \item An unsupervised anomaly detection method for log data, based on an encoder transformer architecture.
    \item An unsupervised decision boundary calculation for the anomaly decision, based on a novel data augmentation method for log data.
    \item An evaluation of the method, based on three different real-world and publicly available datasets, which are BGL, thunderbird, and spirit\footnote{https://www.usenix.org/cfdr-data}.
    \item An evaluation of the method, based on an industry dataset from an IT service provider.
\end{itemize}

This paper is structured as follows. First, related work is presented in \autoref{sec:related_work}. A general framework for anomaly detection on log data as well as a problem description for the final anomaly decision are presented in \autoref{sec:3_ad_logdata}. Preliminaries and implementation details of our method are presented in \autoref{sec:4_method}. Finally, \autoref{sec:5_evaluation} presents our evaluation and \autoref{sec:6_conclusion} is concluding the work.

\section{Related Work}
\label{sec:related_work}
Detecting abnormal events in large-scale systems, indicated by log files, is crucial for creating dependable services. Therefore, log analysis becomes increasingly important for industry and academia ~\cite{qi2020small} and a wide range of different anomaly detection techniques have been developed and discussed in detailed surveys~\cite{chandola2009anomaly, he2016experience}. They utilize different forms of log templates and log embeddings to convert the logs to a machine-readable format. Commonly used anomaly detection methods are support vector machines~\cite{liang2007failure} and principal component analysis~\cite{jolliffe2005principal}. 
In addition, there are rule-based~\cite{breier2015anomaly},
tree-based~\cite{chen2004failure},
statistical~\cite{xu2009detecting},
as well as methods based on clustering~\cite{lou2010mining, cinque2012event, baseman2016relational, LANDAUER201894}. 
Whereas recent anomaly detection methods are mainly designed with neural networks~\cite{zhang2019robust,brown2018recurrent,yin2020improving} and based on encoder architectures~\cite{nicolau2016hybrid,sakurada2014anomaly}. Equally, recent methods utilizing the attention mechanism that is often used in encoder architectures~\cite{vaswani2017attention}.

We classify anomaly detection methods in two different types: supervised and unsupervised~\cite{7774521}. Supervised methods are usually more accurate, though they train the anomalies of the specific dataset as well ~\cite{zhang2019robust, yang2019nlsalog}. However, not all anomalies can be known in advance. Hence, in industrial applications, unsupervised methods are more practical as anomaly labels are mostly unavailable~\cite{meng2019loganomaly, wittkopp2020decentralized}. 

Consequently, we focus on unsupervised methods in this paper. Several unsupervised learning methods, based on neural networks, have been proposed, of which we present a selection in the following.

Du et al.~\cite{du2017deeplog} proposed DeepLog, a Long short-term memory (LSTM) network architecture that is capable of identifying abnormal sequences of log messages. For this, log templates are generated and sequences of templates are formed as model inputs.
The model provides a ranked output with probabilities for the next template in a given sequence.
The anomaly detection is then based on whether the next template has a high probability or not. 

LogAnomaly~\cite{meng2019loganomaly} is similar to DeepLog and predicts the next log message in a sequence of log messages. Instead of utilizing sequences of log templates, LogAnomaly utilizes sequences of log embeddings to improve prediction effectiveness. 

Yang et al.~\cite{yang2021semi} combining the attention mechanism with a gated recurrent network architecture to perform anomaly detection on log data. Thereby the log messages are transformed into log templates, to then predict if a sequence of log templates is normal or abnormal. Furthermore, the labels of the training date are estimated to incorporate knowledge on historical anomalies into the model.

Another approach is Logsy as proposed by Nedelkoski et al.~\cite{nedelkoski2020self}. This approach also incorporates the attention mechanism with an encoder architecture, where log embeddings are calculated. 
The embeddings of normal log messages are condensed into a centroid using a hyperspherical loss function, whereby embeddings of abnormal log messages are pushed away. 
The anomaly detection task is then based on the distance to the centroid

Guo et al.~\cite{guo2021logbert} introduce LogBERT. 
It utilizes the transformer network from BERT~\cite{devlin2018bert}, which consists of an encoder and a decoder, including the attention mechanism. Like DeepLog, it tries to predict a targeted log template of a sequence.
Therefore, they utilize temporal-related log embeddings around the prediction target as inputs.
To predict the targeted log template, it utilizes Cross-Entropy-Loss, extended by a hyperspherical loss function, to ensure compactness of the embeddings.

However, all methods presented above use a manually set decision boundary for anomaly detection. However, optimization requires knowledge about the anomalies, which limits these methods.

\section{Anomaly Detection on Log Data}
\label{sec:3_ad_logdata}
In this section, we describe the general framework for anomaly detection on log data. 
Afterwards, we describe the problem of automatically finding an accurate decision boundary for the anomaly detection task.

\subsection{General Framework}
Anomalies are patterns in data that do not conform to a defined notion of normal behavior~\cite{chandola2009anomaly}.
Anything that deviates from normal behavior can be considered abnormal behavior. 
Steinwall et al.~\cite{steinwart2005classification} state that this can be considered as a binary classification task.
Consequently, anomaly detection on log data is defined as the problem of assigning a binary label to each log message. 
There are two principal approaches to anomaly detection: Supervised and unsupervised. 
In terms of anomaly detection on log data, supervised means that both classes, normal and abnormal, are used during the training phase. 
Unsupervised means that the model is trained on normal log messages only. 
As not all possible anomalies in log data can be known and used for training~\cite{ramponi2020neural}, unsupervised approaches are well suited for log data scenarios, and thus are of high interest for industry and academia~\cite{qi2020small,baier2019cope}. 
Therefore, the challenge is to develop a good understanding of normal log messages, e.g. by internalizing their usual structure, such that any significant deviations can be treated as log messages representing abnormal behavior. 
Thus, a general framework for unsupervised approaches can be described as follows:

\begin{figure}[htbp]
\centering
\includegraphics[width=1\columnwidth]{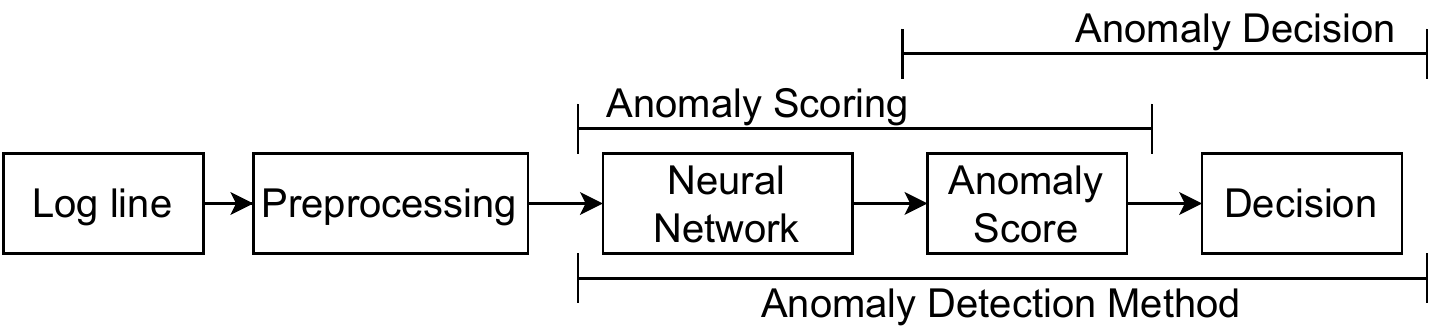}
\caption{Structure of an unsupervised anomaly detection method for log data content.}
\label{fig:ad_method}
\end{figure}

\autoref{fig:ad_method} illustrates the process of the anomaly detection task. 
First, the log message must be transformed into a format that is suitable for machine learning algorithms.
Since anomaly detection models are often designed as neural networks~\cite{naseer2018enhanced,du2017deeplog,nedelkoski2020self} which we also utilize in this work, we consider the case of employing a neural network architecture.
The output of a neural network is commonly transformed such that it can be interpreted as a probability or a probability distribution over one or multiple output neurons, from which further decisions can be inferred.
Therefore, an anomaly detection method requires a final decision on interpreting the neural network's output. 
From this, we conclude that an unsupervised anomaly detection method, which includes a neural network, consists of two parts:
The \textit{Anomaly Scoring} and the \textit{Anomaly Decision}.
The neural network learns the normal behavior in the first part and calculates an anomaly score for each log message.
Therefore, the \textit{Anomaly Scoring} must transform each log message into a real-valued output. 
The second part is the final anomaly detection decision, based on the anomaly scoring, which transforms the scores into a binary classification.
Formally, the anomaly detection model $\Phi$ for log data $L$ is described by two functions, $s$ and $a$, where $s: L \rightarrow \mathbb{R}$ is the \textit{Anomaly Scoring} function, represented by a neural network and $a: \mathbb{R} \rightarrow \{0,1\}$ is the \textit{Anomaly Decision}. Therefore, $\Phi = a(s(x))$, where $x$ is the preprocessed log data input.

\subsection{Problem Description}
\label{sec:problem_definition}

Two main challenges naturally arise when designing an unsupervised anomaly detection method, in our case on log data.
On the one hand, the respective model is purely trained on normal log messages, which makes it difficult to identify abnormal log messages as such, as the computed anomaly scores cannot be interpreted, and thus scores of real abnormal log messages can have an arbitrary form.
On the other hand, the decision boundary for separation of both classes is also configured based on normal log messages only. 
A solution to this problem needs to take these challenges into consideration.

\begin{figure}[htbp]
\centering
\includegraphics[width=\columnwidth]{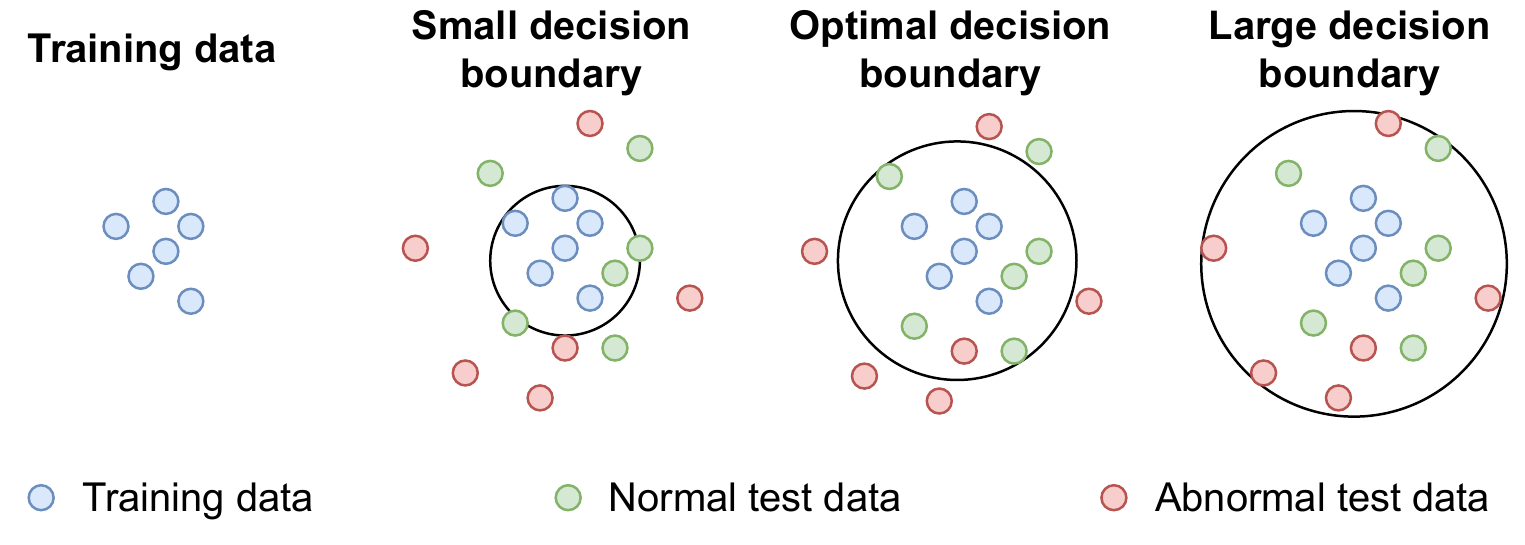}
\caption{Illustration of the various possibilities for manually setting a decision boundary.}
\label{fig:decision_boundary}
\end{figure}
\autoref{fig:decision_boundary} demonstrates the difficulty of setting a precise decision boundary by only utilizing normal data points from the training. 
The blue points represent normal data points, which are part of the training data. 
The green and red points represent normal and abnormal data points during the test phase. 
It can be observed that the green points are on average closer to the normal trained data points, yet considering them for configuration of the decision boundary would lead to a suboptimal separation. 
If the decision boundary is too small, too many log messages will be classified as false positive by the method. 
If the decision boundary is too large, too many log messages will be classified as false negatives. 
Both scenarios result in weaker precision, recall, and F1 scores. 
This example demonstrates that both components, the \textit{Anomaly Scoring} and the \textit{Anomaly Decision}, have to be precise. 

Due to the various factors influencing the training of a neural network, each trained anomaly scoring model will most likely produce different anomaly scores for the same data, and therefore the final decision boundary must be set individually for each model.
Moreover, as the nature and relation of future normal and abnormal log messages cannot be known in advance, it is required that the decision boundary can cope with these uncertainties and still provide good decisions solely based on seen normal log messages.

\section{Method}
\label{sec:4_method}

This section describes in detail our approach A2Log for unsupervised anomaly detection on log data, including our proposed solutions for anomaly scoring and anomaly decision. 
Prior to that, we briefly introduce the necessary preliminaries.

\subsection{Preliminaries}
Logging is commonly employed in order to investigate faulty behavior of systems and services and to increase dependability, which results in information being written to a log file.
The log file documents the executions of the software and is created by log instructions (e.g. \texttt{printf()} or \texttt{log.info()}). Each log instruction results in a single log message, such that the complete log is a sequence of messages $\mathcal{L} = (l_i : i = 1,2,3,\ldots)$. 
There is a commonly used separation in \textit{meta-information} and \textit{content}. The meta-information can contain various information, for example, timestamps or severity levels. The content is free text and consists of a static and a variable part. The static part is called log template.
To access the content of a log message $l_i$, we write $c_i$.

In order to transform the content into a representation that an algorithm can process, methods from the research field of Natural Language Processing (NLP) are applied. Two main concepts are tokenization and embeddings. 

\textbf{Tokenization.} This process splits written text into segments (e.g., words, word stems, or characters).
The smallest indecomposable unit within a log content $c_i$ is a token. 
Consequently, each log content $c_i$ can be interpreted as a sequence of tokens $t_i$
\begin{equation}
t_i = (w_j: w_j \in V, j = 1,2,3,\ldots),    
\end{equation}
where $w$ is a token, $V$ is a set of all known tokens commonly referred to as the \textit{vocabulary}, $j$ is the positional index of a token within the token sequence. To access a specific token at position $j$, we write $t_i^j$. 
Thus, tokenization translates a text into a sequence of tokens using a vocabulary.
The amount of tokens $n_i$ for each log message and the structure of each token can vary, depending on the concrete tokenization method.

\textbf{Embeddings.} 
As the tokens themselves are words of the vocabulary $V$, they cannot be passed into a neural network directly. Furthermore, tokens do not provide any information about their similarity or difference to each other, hence, so called \emph{embeddings} are used to compute a representation of the tokens such that a machine learning model can process it.
Embeddings are real-valued vector representations $\vec{v} \in \mathbb{R}^d$ of either token sequences or a single token; 
a transformation function $g: V \rightarrow \mathbb{R}^{d}$ transforms a token $w_j$ into an embedding $\vec{v_j}$. 
Thereby the same tokens receive the same embeddings.
The sequence of embeddings $\vec{e_i}$, that is describing the corresponding sequence of tokens $t_i$, is defined as follows:
\begin{equation}
\vec{e_i} = (g(w_j): w_j \in t_i, j = 1,2,\ldots,|t_i|),    
\end{equation}
To access the $j$-th embedding in a sequence of embeddings, we write $\vec{e_i}(j)$.
Embeddings are trainable units adapted during the training process to represent the meaning of the underlying token or sequence of tokens.

\subsection{A2Log}
As a solution for an unsupervised anomaly detection method, we propose \textit{A2Log}. A2Log consists of two parts: the \textit{Anomaly Scoring} and the \textit{Anomaly Decision}. To provide a general anomaly detection method, we only utilize the content of log messages, which is a commonality of different log types.

\begin{figure}[htbp]
\centering
\includegraphics[width=\columnwidth]{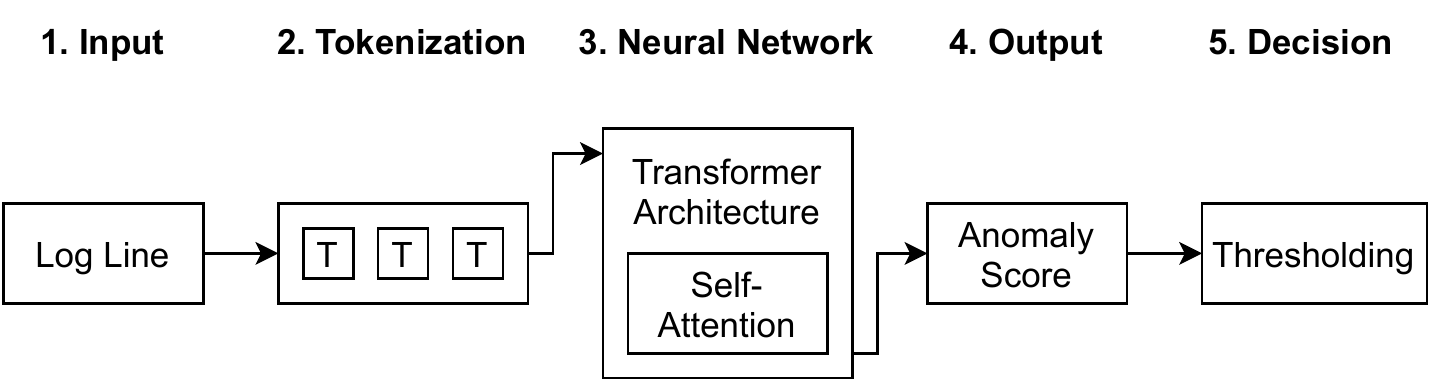}
\caption{Five steps to classify a log message utilizing a transformer architecture.}
\label{fig:steps}
\end{figure}

The schematic flow in~\autoref{fig:steps} of A2Log is as follows.

During the first step, the content $c_i$ of every log message $l_i$ is extracted and in the process tokenized to a sequence of tokens $t_i$ using the symbols \texttt{.,:/} and whitespaces as separators.
Subsequently, we further clean the resulting sequence of tokens by replacing certain tokens with placeholders that adequately represent the original token without losing relevant information. We introduce placeholder tokens for hexadecimal values \texttt{[HEX]} or any number greater or equal 10  \texttt{[NUM]}. 
Finally, we prefix the sequence of transformed tokens with a special placeholder token \texttt{[CLS]} which will be beneficial later on. 
An exemplary log message
\begin{center}
    \texttt{time.c: Detected 3591.142 MHz.}
\end{center}
is thus transformed into a sequence of tokens $t_i$:
\begin{center}
    \texttt{[[CLS], time, c, Detected, [NUM], [NUM], MHz]}.
\end{center}
The token sequence $t_i$ serves as the input for our \textit{Anomaly Scoring} model. Therefore, we utilize the encoder of the transformer architecture ~\cite{devlin2018bert} with self-attention ~\cite{vaswani2017attention}. We have chosen this model architecture since it has already performed well in the domain of natural language processing. The encoder of the transformer architecture is applied to map sequences of tokens onto one d-dimensional vector (embedding), which is represented through the \texttt{[CLS]} token.

\begin{figure}[htbp]
\centering
\includegraphics[width=\columnwidth]{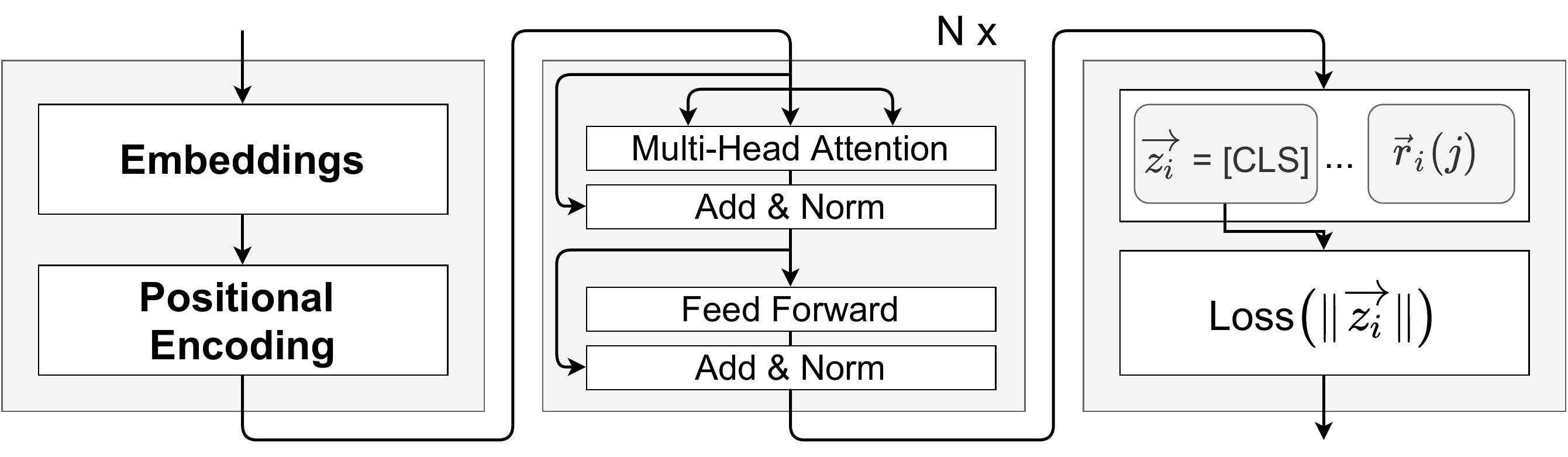}
\caption{Transformer encoder architecture.}
\label{fig:transformer_architecture}
\end{figure}

\autoref{fig:transformer_architecture} depicts the aforementioned network architecture.
During the embedding step, for each token $w_j$ in the token sequence $t_i$, an embedding $\vec{e}_{i}(j)$ is obtained using the transformation function $g$. These embeddings are equal for the same tokens and modified during the training process.
Since these sequences can vary in length, we truncate them to a fixed size $u$ and fill up smaller sequences with padding tokens \texttt{[PAD]}. 
Hence, the employed architecture does not consider the order of the tokens, the input sequence is enriched with positional encoding~\cite{vaswani2017attention,GehringAGYD17}. Based on this, the attention mechanism can take the order of the tokens into account. 

The model then computes an output embedding for each truncated input embedding sequence, which summarizes the log message by utilizing the embeddings of all tokens. 
This output embedding is encoded in the embedding of the \texttt{[CLS]} token and also modified during training, by minimizing the loss. 
During the training process, the model is supposed to learn the meanings of the log messages, thereby getting an intuition of normal log messages.
We denote the output of the model as $\vec{z_i}=s(t_i)$ and use it throughout the remaining steps.
Thereby the anomaly score is calculated by the length of the output vector $\lVert \vec{z_i} \rVert$.
These values must reflect the anomaly probability for each sequence. 
We set the scoring target for normal data to $0$ as the absolute normal state, so that the likeliness of an anomaly in the log message is proportional to increasing positive values. 
The following objective function is utilized to ensure compact anomaly scores near to $\lVert \vec{z_i} \rVert\geq0$ for normal training data~\cite{nedelkoski}:

\begin{equation}
\begin{multlined}
\frac{1}{n}\sum_{i=1}^n(1-y_i)\left\|\vec{z_i}\right\|^2 \\ - y_i \cdot log(1-exp(-\left\|\vec{z_i}\right\|^2))
\end{multlined}
\end{equation}

Thereby, $y_i$ is the label for each input $t_i$, which implies that there must be two classes to train the neural network to avoid the implosion of the model. 
If there is only one class, the objective function will force the model to produce the same result independent of the input. 
Therefore, we define a stabilization class $S$ as the second class which encompasses normal log messages but with different origins. 
This second class is constructed by utilizing normal log messages from other services, where we randomly sample an equal amount of log messages from each service. 
This gives the model an intuition of variety in log messages and, therefore, the model should be able to assign different anomaly scores to real anomalies based on log message characteristics. 
Eventually, the \textit{Anomaly Decision} function assigns each log message an explicit label $\in \{0,1\}$ with

\begin{equation}
a(\left\|\vec{z_i}\right\|)= 
\begin{cases}
    1, & \left\|\vec{z_i}\right\| > \epsilon,\\
    0, & \left\|\vec{z_i}\right\| \leq \epsilon
\end{cases}
\end{equation}
where $\epsilon$ is a decision boundary that is used to obtain binary labels from anomaly scores. 

As we do not know how the \textit{Anomaly Score} will turn out for new normal log messages appearing in the future, we require a method that simulates deviations to the already trained log messages to understand how the model will react, and determines the decision boundary $\epsilon$.
We simulate deviations to the normal training data by applying data augmentation.

\begin{figure}[htbp]
\centering
\includegraphics[width=0.9\columnwidth]{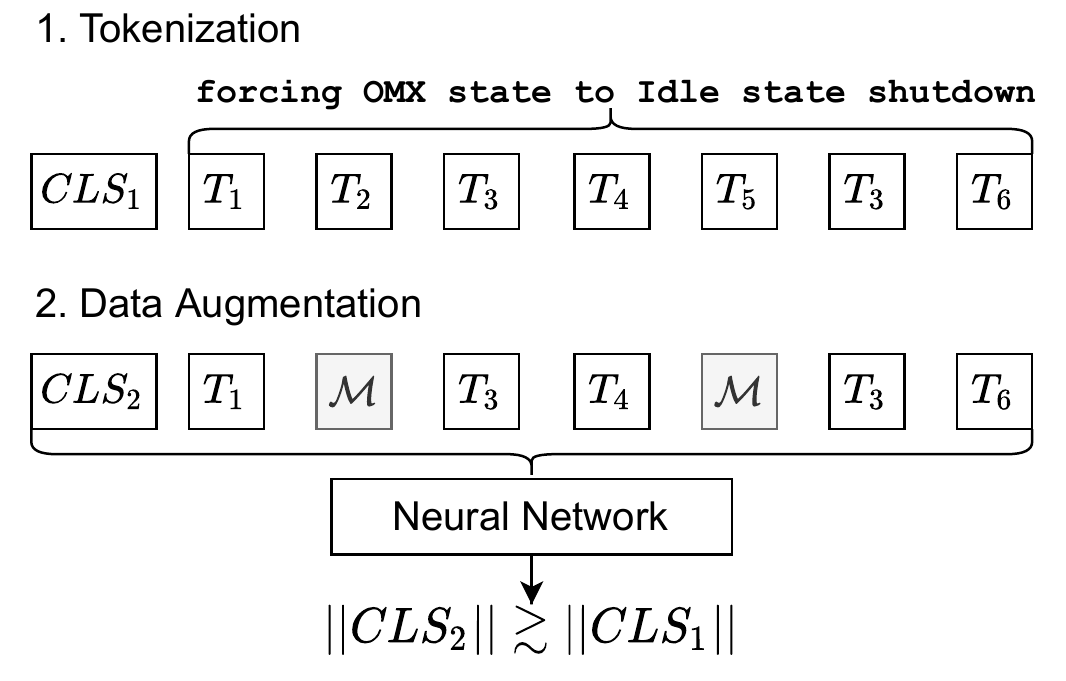}
\caption{Data augmentation with $\alpha=2$ replaced tokens.}
\label{fig:data_augmentation}
\end{figure}

\autoref{fig:data_augmentation} depicts our method for data augmentation on log data. 
The data augmentation is applied to the log messages of the original training dataset $l_{train}$ only, without the stabilization class $S$.
First, the log messages are tokenized, followed by replacing $\alpha \in \mathbb{N}$ tokens at random positions in each token sequence $t_i$ with a masking token \texttt{[$\mathcal{M}$]}, described in \autoref{eq:data_augmentation}.

\begin{equation}
\label{eq:data_augmentation}
t'_i = (w_j : w_j \in V \lor w_j = [\mathcal{M}], j=1,2,\ldots,|t_i|)
\end{equation}
Then the trained model for the \textit{Anomaly Scoring} $s$ calculates a scalar value for each augmented token sequence $t'_i$ which should be greater or equal than the scalar value for the original token sequence $s(t'_i) \gtrsim s(t_i)$ because it contains $\alpha$ unknown tokens \texttt{[$\mathcal{M}$]} and, therefore, does not comply with established knowledge. 
We gather all values - calculated by the model - for the augmented token sequences to receive a distribution $D$. 
\begin{equation}
D = (s(t'_i) : t'_i \in l_{train}, 0 \leq i \leq |l_{train}|).
\end{equation}

This distribution is the basis for the decision boundary $\epsilon$. We calculate the decision boundary by choosing the i-th percentile $p$ of the distribution $D$ and multiplying this value with a regulator variable $\beta$.

\begin{equation}
\epsilon(D, p,\beta) = D_p \cdot \beta
\end{equation}

This formula calculates the final decision boundary from the given distribution for the deviated token sequences. The parameter $p$ is a bias regulator, because there is a possibility, that the model calculates outliers for few normal augmented log messages. These outliers can occur when rare normal log messages resemble the stabilization class data and have high anomaly scores. With $\beta$, we can control whether we want to allow more deviations for normal data during the prediction phase.

\section{Evaluation}
\label{sec:5_evaluation}

To quantify the performance of A2Log, we test it on three real-world HPC log datasets and one industry dataset. Thereby, we compare our anomaly detection method with three other approaches and evaluate the performance of our anomaly decision boundary.

\subsection{Experimental Setup}
For the evaluation, we investigate four different training splits where we use the first 10\%, 20\%, 40\%, 60\% of normal data to train the anomaly scoring model and calculate the anomaly decision boundary. The remaining log messages are used for the testing. For each split, we examine the F1 score, precision, and recall. This setup is applied to the three publicly available datasets as well as to the industry dataset.
\def\arraystretch{1.2}%
{\renewcommand{\arraystretch}{1.2}}
\begin{center}
\begin{table*}[h]
    \centering
    \begin{tabular}{|c|c|c|c|p{1.0cm}|p{1.0cm}|p{1.0cm}|p{1.0cm}|} 
    
    \hline
    & & & & \multicolumn{4}{c|}{\textbf{\#unique log templates in test}} \\
    \textbf{System} & \textbf{\#normal} & \textbf{\#anomalies} & \textbf{\#templates} & \multicolumn{4}{c|}{\textbf{that do not appear in train for splits}}\\\cline{5-8}
    
    & & & & 10 \% & 20 \% & 40 \% & 60 \% \\ 
    \hline
    BGL         & 4,399,503 & 348,460 & 1,571 & 1,318 & 1,232 & 1,158 & 1,077 \\ 
    Thunderbird & 4,773,713 & 226,287 & 1,302 & 232  & 205  & 128  & 61   \\ 
    Spirit      & 4,235,110 & 764,890 & 1,457 & 1,091 & 1,028 & 297  & 129  \\
    \hline
    
    \end{tabular}
    \caption{Dataset description of BGL, Thunderbird and Spirit.}
    \label{tab:datasets}
\end{table*}
\end{center}
\textbf{Publicly Available Datasets.}
We select three real-world datasets from HPC systems for evaluation as target systems, namely Blue Gene/L (BGL), Spirit, and Thunderbird (Tbird)~\cite{oliner2007supercomputers}. From every dataset\footnote{https://www.usenix.org/cfdr-data}, we utilize the first 5 million log messages. 

To reveal the characteristics of the datasets, we have calculated the log templates for each dataset using Drain~\cite{he2017drain}. \autoref{tab:datasets} depicts the count of normal and abnormal data, the count of log templates for the whole dataset, and the number of distinctive log templates that are present in the test dataset only, without being in the training dataset. It can be seen that especially for small amounts of training data, there are a lot of log messages in the test data, that are not seen in the training data, especially on the BGL dataset. 
That makes anomaly prediction difficult when only the normal data is trained, and the normal data in the test set deviates from it. Overall, 7 \% to 15 \% of the log lines are abnormal.

\def\arraystretch{1.2}%
{\renewcommand{\arraystretch}{1.2}}
\begin{table*}[t]
    \centering
    \begin{tabular}{|c|c c|c c c c|}
    \hline
         \parbox{1cm}{\textbf{Train\\split}} & \textbf{\#train} & \parbox{1.4cm}{\textbf{\#unique \\train}} & \textbf{\#test} & \parbox{1.4cm}{\textbf{\#unique \\ test}} & \parbox{1.4cm}{\textbf{\#unique\\test\\abnormal}} & \parbox{1.4cm}{\textbf{\#test \\ abnormal}} \\
         \hline
         0.1 & 711,841 & 226,375 & 6,409,147 & 1,270,089 & 28 & 196,534 \\
         0.2 & 1,424,198 & 465,666 & 5,696,790 & 106,0947 & 27 & 196,519 \\
         0.4 & 2,848,395 & 536,947 & 4,272,593 & 987,417 & 23 & 146,099 \\
         0.6 & 4,272,593 & 895,520 & 2,848,395 & 614,089 & 18 & 100,827 \\
         \hline
    \end{tabular}
    \caption{Dataset description of the industry dataset.}
    \label{tab:meta_data_industry}
\end{table*}

\textbf{Industry Dataset.}
Furthermore, we investigate \textit{A2Log} for an industry dataset that comes from the production environment of an IT service and cloud provider.
Thereby the focus is to ensure a dependable storage service, by detecting anomalies in the log data from the underlying hardware. 
For this, we used log data from different disk controllers that are managing a variety of hard disks.
The task is to identify anomalies by only training normal log data from some hard disks with no recorded anomalies and apply the model to new disks without manual optimization.

\autoref{tab:meta_data_industry} shows the total number of training samples and the number of unique training samples, depending on the split. It also reveals the numbers of total and unique normal samples in the test data. Furthermore, it reveals the count of abnormal samples and the number of unique anomalies in the test dataset. It can be concluded that there are only very few different anomalies in the test dataset but they occur frequently. Furthermore, the rate of anomalies for the dataset is between 3 to 3.5 \%, regarding the respective split. The dataset has 7,120,988 log lines in total.

\textbf{Benchmark Approaches.}
In total, we evaluate four different methods, including \textit{A2Log}. The first baseline is DeepLog, where we set the final decision boundary as best as possible. For the other three methods, we train our described transformer model for the anomaly scoring and then calculate the decision boundary differently. First, we calculate the decision boundary based on the 3-sigma method. The 3-sigma decision boundary is calculated on the anomaly scores of the trained model for all training data, excluding the stabilization class. The second benchmark \textit{Best} is the best possible result for the transformer model, which is achieved through utilizing the test data, including the abnormal samples, to calculate the optimal decision boundary. Therefore, this benchmark cannot be exceeded and is considered as an absolute upper limit. Therefore, the goal is to get as close as possible to this result. We then evaluate \textit{A2Log} against these three benchmarks.

\subsection{Implementation Details}
First, we tokenize the content of each log line as described in \autoref{sec:4_method} and then truncate all sequences of tokens to the same length $|t_i| = 20$. We set the dimensionality $d$ of the embeddings to 128. The weights of the embeddings are initialized with \textit{Xavier}~\cite{glorot2010understanding}. For the anomaly scoring model training, we use a hidden size of $256$, a batch size of $1024$, a dropout rate of $5 \cdot 10^{-2}$. For the optimization task, we use the Adam optimizer with a learning rate of $10^{-4}$ and a weight decay of $5 \cdot 10^{-5}$ in every experiment. 

Since we need a stabilization class $S$ to train our model, we use 60k random messages from other datasets. The corresponding mappings are displayed in \autoref{tab:stabilization}. 

\begin{table}[!h]
    \resizebox{\columnwidth}{!}{%
    \centering
    \begin{tabular}{|c|p{1.0cm}|p{1.0cm}|p{1.0cm}|p{1.0cm}|}
    \hline
    \textbf{Dataset} & \multicolumn{3}{c|}{\textbf{\#logs for stabilization from}} & \\\cline{2-4}
    \textbf{for evaluation} & BGL & Tbird & Spirit & \textbf{\#total}\\
    \hline
    BGL & / & 60000 & 60000 & 120000\\
    Tbird & 60000 & / & 60000 & 120000\\
    Spirit & 60000 & 60000 & / & 120000\\
    Industry dataset & 60000 & 60000 & 60000 & 180000\\
    \hline
    \end{tabular}}
    \caption{Log messages utilized for the stabilization class.}
    \label{tab:stabilization}
\end{table}

\begin{figure*}[h]
\centering
\includegraphics[width=1\textwidth]{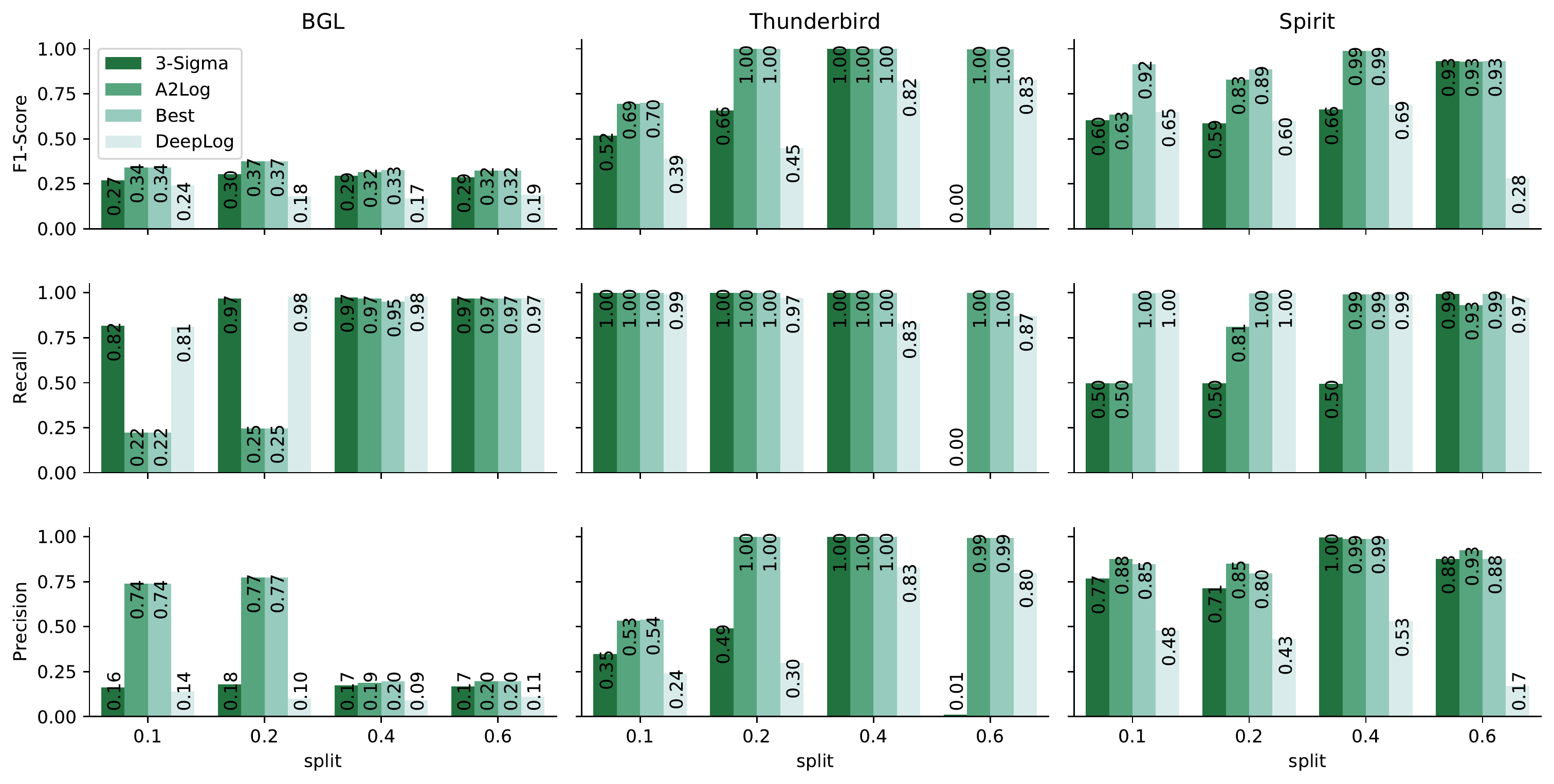}
\caption{Evaluation on BGL, thunderbird and spirit.}
\label{fig:results}
\end{figure*}

\begin{figure*}[h]
\centering
\includegraphics[width=1\textwidth]{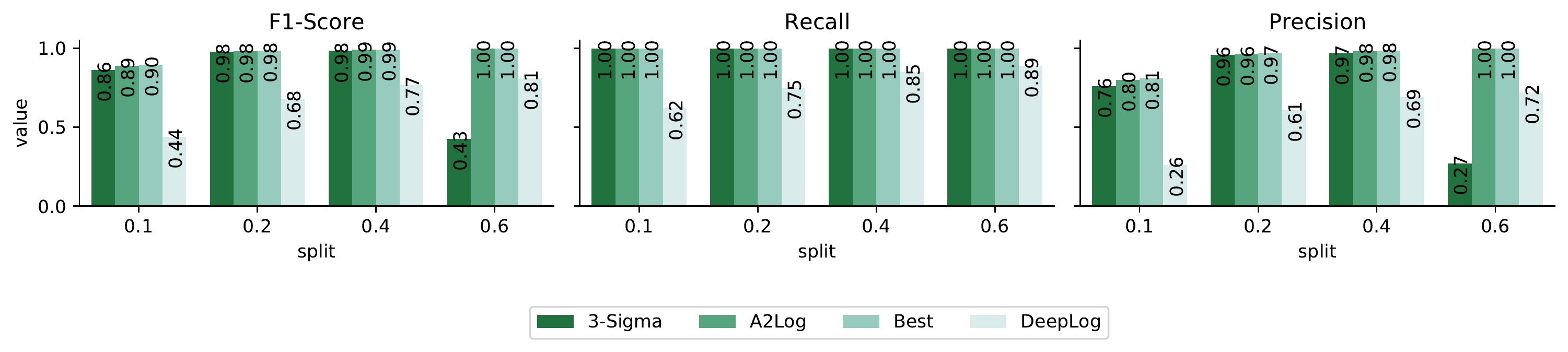}
\caption{Evaluation on the industry dataset.}
\label{fig:results_id}
\end{figure*}

For the three datasets (BGL, Thunderbird, and Spirit), we utilize 120,000 log lines each as a stabilization class. For the industry dataset, we utilize a total of 180,000 log lines as the stabilization class.
Due to the disparity in the number of training samples of the normal log messages and the stabilization class, we use a weighted sampler for the training to balance the training. All models are trained until an average loss of $0.01$ per sample, or a maximum of 50 epochs, is reached.

The anomaly decision function $a$ is parameterized as shown in \autoref{tab:ad_param}.

\begin{table}[!h]
    \resizebox{\columnwidth}{!}{%
    \centering
    \begin{tabular}{|c|c|c|c|c|}
    \hline
         Parameter & BGL & Thunderbird & Spirit & \parbox{1.2cm}{\vspace{.2\baselineskip}industry\\dataset\vspace{.2\baselineskip}}  \\
         \hline
         $\alpha$ & 1 & 1 & 1 & 1 \\
         p & 0.95 & 0.95 & 0.95 & 0.95 \\
         $\beta$ & 2.5 & 2.5 & 5.0 & 2.0 \\
         \hline
    \end{tabular}}
    \caption{Decision boundary parameters to calculate $\epsilon$.}
    \label{tab:ad_param}
\end{table}

We set $\alpha$ to $1$ in each experiment for each dataset to replace only one token of the log message with the unknown token for the data augmentation. Likewise, in each experiment, we set $p$ to 0.95 to filter out 5 \% of the augmented data with the highest anomaly scores. Furthermore, we adjust $\beta$ for each dataset but keep it the same for each experiment on the respective dataset.

\subsection{Results}
For all evaluations, we do every experiment three times and depict the best results in terms of the F1-score in \autoref{fig:results} for the three publicly available datasets and in \autoref{fig:results_id} for the industry dataset.

\autoref{fig:results} depicts the results of \textit{Best}, 3-Sigma and Deeplog, compared with \textit{A2Log}. It shows the respective approaches for different amounts of training data of 10\%, 20\%, 40\%, 60\% of the respective dataset. It can be seen that A2Log is superior or equal to DeepLog and the baseline 3-Sigma in all experiments. In addition, A2Log achieves almost the same F1 scores as \textit{Best} in all experiments. 
A noticeable aspect of the Spirit dataset is that as soon as only a few learning samples are available, the decision boundary function of A2Log no longer turns out to be that precise.
Nevertheless, the other training splits for the Spirit dataset show that our unsupervised A2Log approach can perform equally to an optimal decision boundary. 
The results on the BGL dataset are not particularly high, as there is a fundamental concept drift in the data, which can be seen by the fact that even for high training splits of 60 \% there are still many log templates present in the test dataset that are not present in the training dataset as shown in \autoref{tab:datasets}.

In addition, the two parameters $\alpha$ and $p$ can be set to 1 and 0.95 independently of the dataset. This shows that the method is robust in parameter choice and independent of the dataset.

\autoref{fig:results_id} depicts the performance of \textit{A2Log} for different training splits. Thereby it compares A2Log with the best possible decision boundary and the 3-sigma boundary. It can be observed that \textit{A2Log} outperforms 3-Sigma and is as good as the best possible decision boundary.
When \textit{A2Log} can train enough data, it is able to identify anomalies on new hard disks with perfect scores.

\section{Conclusion}
\label{sec:6_conclusion}
Anomaly detection methods have become increasingly important to ensure the dependable and stable operation of IT services, including their serviceability.
However, existing unsupervised anomaly detection methods are applied under constrained assumptions for the final anomaly decision.
Therefore we propose A2Log, to address the current limitations of unsupervised anomaly detection methods.
Unlike other unsupervised approaches, it calculates its decision boundary for the final decision by exploring the model behavior based on augmented training data. 
With data augmentation, we simulate deviations in log data that occur from service updates over time.
We evaluate our approach on three publicly available datasets and one industry dataset. Thereby, we show the effectiveness of our approach when it is applied in an unsupervised setting.

Even if we can simulate deviations of the normal log lines with the help of our data augmentation and thus classify new normal log lines as such, we see as a limitation that a fundamental concept drift in the data pushes the method to its limits and thus leads to misclassifications.
Nevertheless, A2Log can keep up with the optimal decision boundary, which can only be calculated by utilizing available anomaly examples. 
Furthermore, we are able to show that the entire method outperforms DeepLog and 3-Sigma as a baseline decision boundary.

As further work, we consider additional adjustments to the data augmentation process, to integrate it into other anomaly detection methods. Thereby, we want to apply data augmentation techniques that are well known in the research area of natural language processing. Furthermore, we want to extend A2Log to other anomaly detection settings, e.g. supervised or weakly supervised settings. 

\bibliographystyle{ieeetr}
\bibliography{sample}

\end{document}